\documentclass[letterpaper, 10 pt, conference]{ieeeconf}  

\IEEEoverridecommandlockouts                              

\usepackage{graphics} 
\usepackage{epsfig} 
\usepackage{times} 
\usepackage{amsmath} 
\usepackage{amssymb}  

\usepackage{ulem}       
\usepackage{subfig}     
\usepackage{booktabs}   
\usepackage{multirow}   
\usepackage[hidelinks]{hyperref}   
\usepackage{tabularx}   
\usepackage[table]{xcolor}     

\newlength\savewidth
\newcommand\shline{\noalign{\global\savewidth\arrayrulewidth
  \global\arrayrulewidth 1pt}\hline\noalign{\global\arrayrulewidth\savewidth}}

\title{\LARGE \bf
DSLO: Deep Sequence LiDAR Odometry Based on Inconsistent Spatio-temporal Propagation}

\author{Huixin Zhang\textsuperscript{\dag1}, Guangming Wang\textsuperscript{\dag2}, Xinrui Wu\textsuperscript{1}, Chenfeng Xu\textsuperscript{3}, \\ Mingyu Ding\textsuperscript{3}, Masayoshi Tomizuka\textsuperscript{3}, Wei Zhan\textsuperscript{3}, and Hesheng Wang\textsuperscript{1} 
\thanks{\textsuperscript{\dag}The first two authors contributed equally.}
\thanks{This work was supported in part by the Natural Science Foundation of China under Grant 62073222, U21A20480 and U1913204, in part by the Science and Technology Commission of Shanghai Municipality under
Grant 21511101900, in part by the Open Research Projects of Zhejiang Lab under Grant 2022NB0AB01. 
Corresponding Author: Hesheng Wang. 
}
\thanks{\textsuperscript{1}Department of Automation,
Key Laboratory of System Control and Information Processing of Ministry of
Education, Key Laboratory of Marine Intelligent Equipment and System of
Ministry of Education, Shanghai Engineering Research Center of Intelligent
Control and Management, Shanghai Jiao Tong University, Shanghai 200240,
China. }
\thanks{\textsuperscript{2}Department of Engineering, University of Cambridge, Cambridge CB2 1PZ, U.K. }
\thanks{\textsuperscript{3}UC Berkeley, Berkeley, CA 94720 USA.}
}

\begin{document}
\maketitle
\thispagestyle{empty}
\pagestyle{empty}
	
\begin{abstract}

This paper introduces a 3D point cloud sequence learning model based on inconsistent spatio-temporal propagation for LiDAR odometry, termed DSLO. It consists of a pyramid structure with a spatial information reuse strategy, a sequential pose initialization module, a gated hierarchical pose refinement module, and a temporal feature propagation module. 
First, spatial features are encoded using a point feature pyramid, with features reused in successive pose estimations to reduce computational overhead. 
Second, a sequential pose initialization method is introduced, leveraging the high-frequency sampling characteristic of LiDAR to initialize the LiDAR pose. 
Then, a gated hierarchical pose refinement mechanism refines poses from coarse to fine by selectively retaining or discarding motion information from different layers based on gate estimations. 
Finally, temporal feature propagation is proposed to incorporate the historical motion information from point cloud sequences, and address the spatial inconsistency issue when transmitting motion information embedded in point clouds between frames. 
Experimental results on the KITTI odometry dataset and Argoverse dataset demonstrate that DSLO outperforms state-of-the-art methods, achieving at least a 15.67\% improvement on RTE and a 12.64\% improvement on RRE, while also achieving a 34.69\% reduction in runtime compared to baseline methods. Our implementation will be available at \url{https://github.com/IRMVLab/DSLO}.


\end{abstract}

\section{Introduction}

LiDAR odometry is a pivotal task in the realm of autonomous navigation~\cite{li2023hong, li2020robust}.
Over the past decade, traditional geometry-based methods have been the cornerstone of LiDAR odometry, providing robust interpretability and operational efficiency~\cite{LOAM, GICP, LeGO-LOAM, suma}. 
However, ideal assumptions in traditional methods can lead to inaccurate system modeling. With advancements in computational hardware, focus has shifted towards leveraging deep learning techniques to tackle LiDAR odometry challenges. Recent works~\cite{LodoNet, LONET, liu2023translo, liu2023regformer, PWCLO, zhou2023hpplo} have explored learning deep feature representations or directly estimating vehicle motion through end-to-end training.

However, learning-based LiDAR odometry faces three main challenges:
1) Projecting unstructured point clouds onto 2D planes~\cite{LodoNet, LONET, liu2023translo} results in loss of 3D spatial information and efficiency.
2) Coarse-to-fine optimization~\cite{PWCLO, zhou2023hpplo} fails to account for varying information reliability across scales.
3) Most methods predict poses using only two adjacent frames, ignoring historical motion information.

\begin{figure}[t]
    \centerline{\includegraphics[width=1\linewidth]{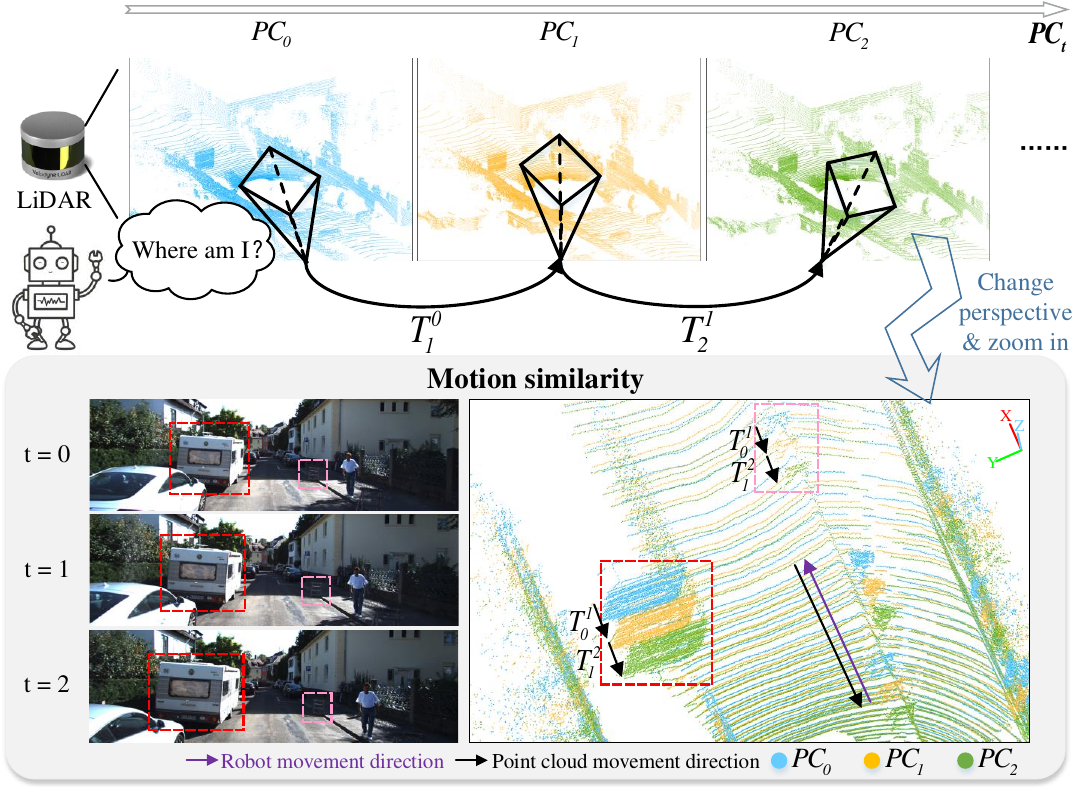}}
    \vspace{-4pt}
    \caption{
    Inspiration for our work. 
    Boxes of the same color indicate the same rigid object. The robot's motion can be inferred from these objects, showing high similarity between adjacent frames.
    }
    \vspace{-15pt}
    \label{fig-pose}
\end{figure}

\begin{figure*}[ht]
    \centerline{\includegraphics[width=1.0\linewidth]{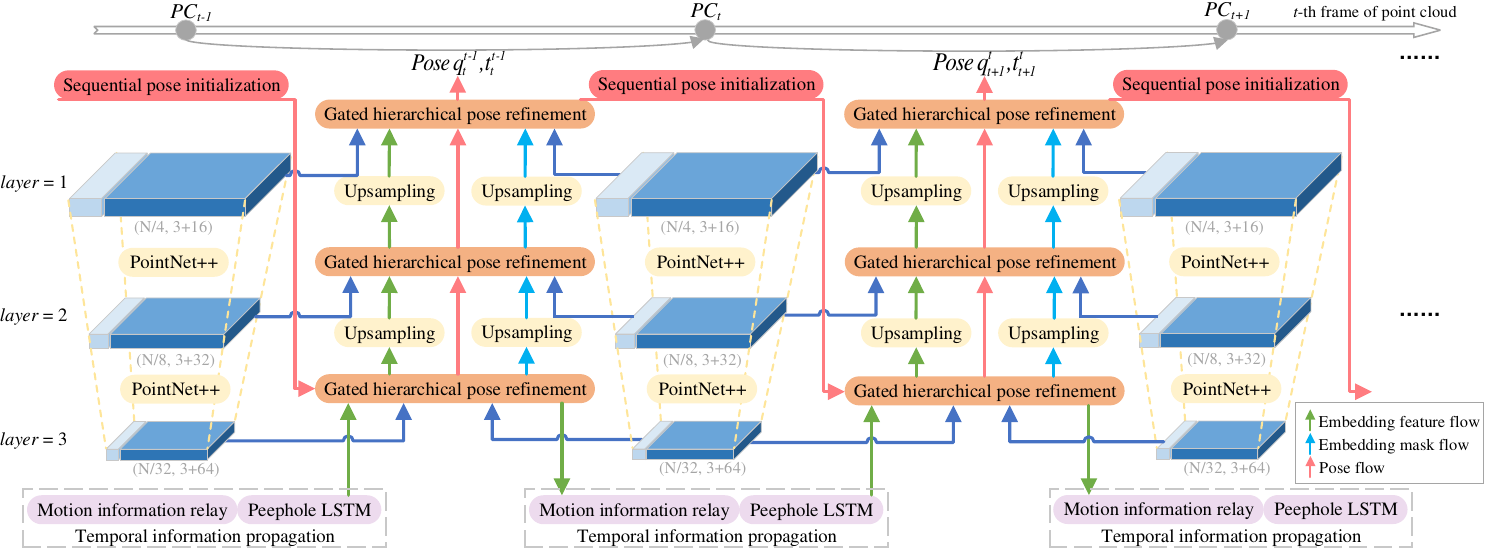}}
    \vspace{-1pt}
    \caption{Overview of our DSLO.
    For pose estimation between two adjacent frames, we encode the point feature pyramid and use a gated hierarchical pose refinement module to achieve coarse-to-fine update.
    For multiple frames,
    we reuse the feature pyramid and 
    utilize the last refined pose as the current initial guess.
    Temporal feature propagation fuses motion features along time series and addresses the spatial inconsistency of point-wise features between frames.
    }
    \vspace{-10pt}
    \label{fig-model}
\end{figure*}

To address these, we propose DSLO, an end-to-end deep sequence LiDAR odometry method leveraging inconsistent spatio-temporal propagation.
Compared to the baseline method PWCLO-Net~\cite{PWCLO}, 
DSLO accelerates inference with spatial information reuse and sequential pose initialization modules.
Gated hierarchical pose refinement is proposed to update features from coarse to fine,
using self-learning gate estimations to mitigate incorrect historical matches in the upper layer and sensor noise in the current layer.
Furthermore, we explore temporal information fusion in LiDAR odometry for better pose regression.
The challenge of modeling temporal motion information in unstructured point clouds and propagating them within inconsistent spatial contexts is addressed. 
The contributions are as follows:
\begin{itemize}
    \item 
    Sequential pose initialization is proposed to reduce computation overhead
    while retaining motion similarity.

    \item Gated hierarchical pose refinement is proposed, 
    which utilizes multi-scale spatial information for hierarchical updates and employs self-learning gate estimations to filter valid information from different layers.

    \item Temporal feature propagation
    is proposed to fuse information across time series.
    Motion information relay is designed to address the spatial inconsistency of the point-wise motion information between frames.

    \item 
    Our method is validated on KITTI~\cite{KITTI1, KITTI2} and Argoverse~\cite{argo} datasets.
    It outperforms state-of-the-art learning-based LiDAR odometry approaches and even some geometry-based methods, while achieving real-time performance on consumer-grade GPUs.
\end{itemize}


\section{Related Work}

\subsection{Deep LiDAR Odometry}



Deep LiDAR odometry often preprocesses point cloud data due to its sparse and disordered nature. LodoNet~\cite{LodoNet} uses spherical projection to create depth images and applies SIFT for keypoint matching. LO-Net~\cite{LONET} projects the point cloud onto a 2D plane and uses normal vector similarity for geometric consistency. DMLO~\cite{DMLO} employs cylindrical projection, CNN-predicted point pairs, and singular value decomposition to obtain the rigid transformation.

End-to-end deep 3D LiDAR odometry uses flow embedding to describe global motion between point clouds, circumventing point-to-point matching errors. DeepCLR~\cite{DeepCLR} estimates flow embedding with MLP and uses FC layers for pose prediction. PWCLO-Net~\cite{PWCLO} predicts flow embedding via attentive cost volume and updates poses in a coarse-to-fine manner.


\subsection{Spatio-temporal Fusion on 3D Point Cloud Learning} 
Several 3D point cloud learning models focus on spatio-temporal information fusion. ASTA3DConv~\cite{wang2021anchor} introduces a novel spatio-temporal convolution for dynamic 3D point cloud sequences, using spatio-temporal attention in neighborhood aggregation around virtual anchors. A self-supervised 4D convolution neural network~\cite{wang2021self} predicts the temporal order of point cloud clips to learn 4D spatio-temporal features, evaluated on nearest neighbor retrieval and action recognition tasks. Similarly, P4Transformer~\cite{fan2021point} uses point 4D convolution to encode and aggregate spatio-temporal features from point cloud videos. BE-STI~\cite{wang2022sti} predicts class-agnostic motion with bidirectional enhancement of spatio-temporal features, leveraging similarities and differences between consecutive and nonadjacent frames.

\section{Methodology}

Fig.~\ref{fig-model} illustrates the overall structure of the proposed network DSLO. 
Firstly, spatial information reuse is introduced to reduce the computational overhead in Sec.~\ref{sec:sharing}.
Secondly, we propose the sequential pose initialization based on motion similarity during two LiDAR sampling intervals in Sec.~\ref{sec:Initialization}.
Next, motion features and poses are updated from coarse to fine through a gated hierarchical pose refinement module in Sec.~\ref{sec:warp}.
Finally, temporal information propagation is proposed in Sec.~\ref{sec:LSTM}, with motion information relay solving the spatial inconsistency of point-wise motion information between frames. The training loss formulation is derived in Sec.~\ref{sec:loss}.
\begin{figure}[t]
	\begin{center}
		\includegraphics[width=0.98\linewidth]{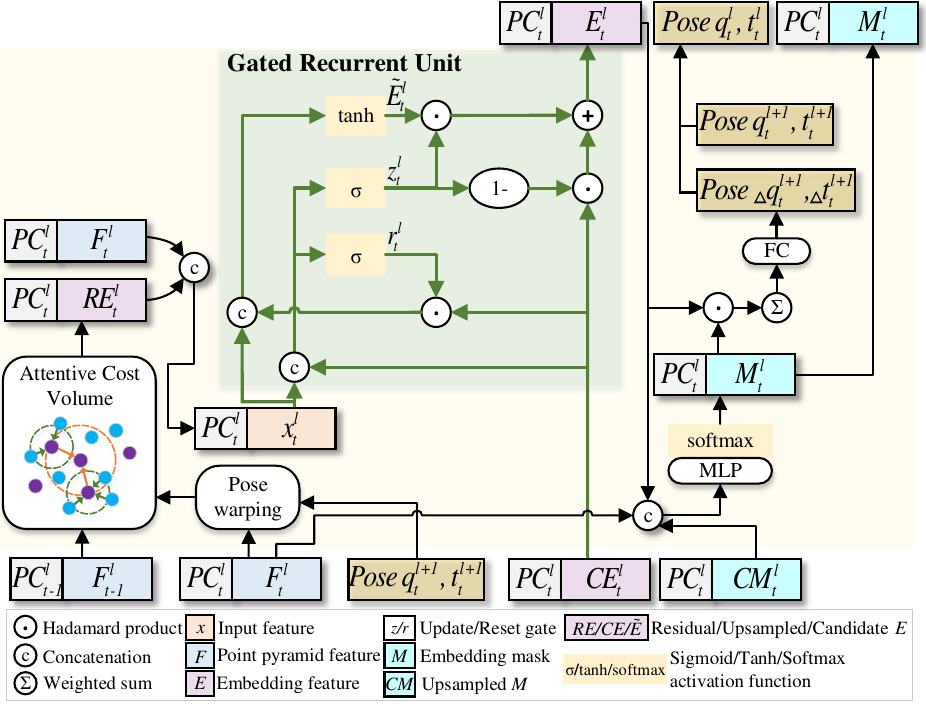}
	\end{center}
	\vspace{-10pt}
	\caption{
        Gated hierarchical pose refinement module. 
        The residual embedding feature ${RE}_t^l$, point pyramid feature $F_t^l$ and upsampled embedding feature $CE^{l}$ are encoded and fed into a GRU. Subsequently, the output embedding feature ${E}_t^l$ assists in refining embedding mask ${M}_t^l$ and pose ${q}_t^l$, ${t}_t^l$.
	}
    \vspace{-10pt}
    \label{fig-warp}
\end{figure}
\subsection{Spatial Information Reuse}\label{sec:sharing}
We construct a point feature pyramid based on PointNet++~\cite{POINTNET2} to extract multi-scale spatial structure information from point clouds.
However, one point cloud is used twice for pose estimations of point cloud sequences. 
The redundant feature extraction of the same point cloud is unnecessary.
To address this inefficiency, we introduce the spatial information reuse strategy.
As depicted in Fig.~\ref{fig-model}, pyramid features of $PC_t$ are stored and reused in two consecutive pose estimations, eliminating redundant operations of downsampling and neighborhood feature aggregation. This strategy contributes to reducing time consumption and computational overhead.

\subsection{Sequential Pose Initialization}\label{sec:Initialization}
As LiDAR is capable of collecting point clouds at a high frequency of 10-100Hz, the pose of LiDAR exhibits minimal variation in two adjacent sampling moments.
Inspired by this, we propose a sequential pose initialization strategy to utilize the motion similarity and improve the real-time performance.

When performing the multi-frame LiDAR odometry task on a point cloud sequence,
we employ the attentive cost volume~\cite{PWCLO} method to estimate the pose between the first two frames. Subsequently, we take the last refined pose as the initial value for the current estimation based on the motion similarity.
Then the initialized pose acts as a prior and is further refined.
As illustrated in Fig.~\ref{fig-model}, the refined pose $q_t^{t-1}$ and $t_t^{t-1}$ between the ($t$-$1$)-th and the $t$-th frame serve as the initial guess for the subsequent estimation between the $t$-th and ($t$+$1$)-th frames of the point cloud.
This sequential pose initialization strategy is iteratively applied throughout the entire sequence.

\begin{figure*}[t]
    \begin{center}
        \includegraphics[width=0.9\linewidth]{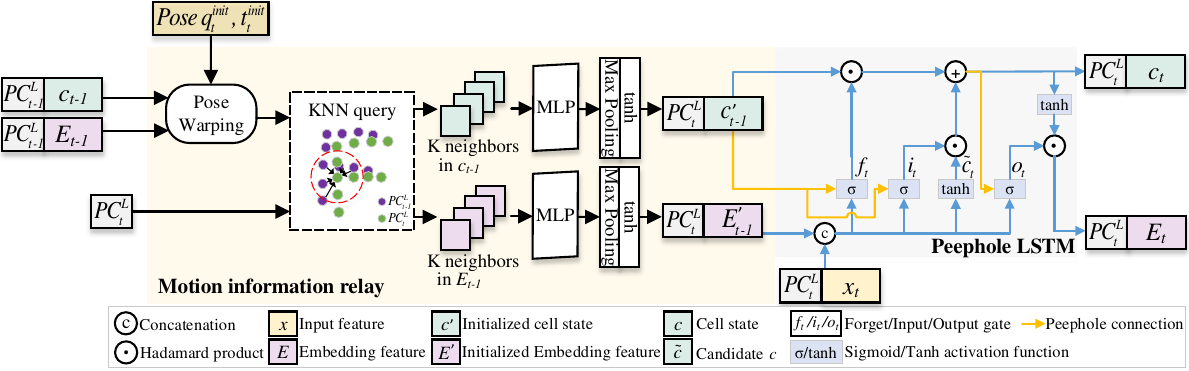}
    \end{center}
    \vspace{-10pt}
    \caption{Temporal feature propagation with inconsistent spatial context.
    The historical motion information $E_{t-1}$ and LSTM cell state $c_{t-1}$ embedded in ${PC}_{t-1}^L$ are passed to $PC_t^L$ with a learning-based motion information relay method. Then peephole LSTM is employed to fuse the temporal motion information.
    }
    \vspace{-10pt}
    \label{fig-transfer}
\end{figure*}

\subsection{Gated Hierarchical Pose Refinement}\label{sec:warp}
We propose a gated hierarchical pose refinement module that leverages multi-scale spatial information and self-learning gate estimations to selectively discard erroneous motion information from different layers.


Fig.~\ref{fig-warp} illustrates the module, which uses upsampling layers~\cite{FLOWNET} to connect different layers. 
The pose in the upper layer is used to warp ${PC}_t^l$ and the residual embedding feature ${RE}_t^{l}$ is calculated by attentive cost volume~\cite{PWCLO} to estimate residual motion.
Then, a Gated Recurrent Unit (GRU)~\cite{FLOWSTEP,RAFT} structure hierarchically updates the embedding feature. The residual embedding feature ${RE}_t^l$, point pyramid feature $F_t^l$ and upsampled embedding feature $CE_t^{l}$ are fed into GRU to obtain the refined embedding feature $E_t^l$ as follows:
{
\setlength{\abovedisplayskip}{2pt}
\setlength{\belowdisplayskip}{2pt}
\begin{align}
\centering
    x_t^{l}&={RE}_t^l \oplus F_t^l, \nonumber
\end{align}
\begin{align}
\centering
    z_t^{l}&=\sigma\left(MLP_z\left({CE}_t^{l}\oplus x_t^{l}\right)\right), \nonumber
\end{align}
\begin{align}
\centering
    r_t^{l}&=\sigma\left(MLP_r\left({CE}_t^{l}\oplus x_t^{l}\right)\right),  \nonumber
\end{align}
\begin{align}
\centering
    \tilde{E}_t^{l}&=\tanh \left(MLP_E\left(r_t^{l} \odot {CE}_t^{l}\oplus x_t^{l}\right)\right), \nonumber
\end{align}
\begin{align}
\centering
    E_t^{l}&=\left(1-z_t^{l}\right) \odot {CE}_t^{l}+z_t^{l} \odot \tilde{E}_t^{l},
    \vspace{-10pt}
    \label{input}
\end{align}
}where $\oplus$ denotes vector concatenation in the feature dimension and $\odot$ denotes the Hadamard product. $\sigma(\cdot)$ denotes the sigmoid activation function. 

The two learnable gates in the gated hierarchical pose refinement module, reset gate $r_t^l$ and update gate $z_t^l$ estimate the confidence weights of motion information from the upper layer and the current layer.
A small activation weight of reset gate $r_t^l$ prevents unreliable point correspondences in the coarse upper layer, while a small activation weight of update gate $z_t^l$ eliminates outliers in the dense point cloud caused by occlusion or sensor noise in the current layer.

Finally, the refinement of the embedding mask $M_t^{l}$ is facilitated by the combined contributions of $E_t^l$, $F_t^l$, and the upsampled embedding mask ${CM}_t^{l}$. It is noted that in the 3-rd layer, $M_t^{l}$ is predicted solely by $E_t^l$ and $F_t^l$ since there is no higher layer.
The pose is then updated by inputting the weighted sum of $E_t^{l}$ and $M_t^{l}$ into an FC layer following~\cite{PWCLO}.

\subsection{Temporal Feature Propagation with Inconsistent Spatial Context}\label{sec:LSTM}
We propose a novel temporal feature propagation module to fuse motion features along time series, so that pose estimations can benefit from historical motion information and local constraints in the time domain.

The structure of our temporal feature propagation module is depicted in Fig.~\ref{fig-transfer}.
A peephole LSTM~\cite{peephole1, peephole2} is introduced for the temporal information propagation.
However, 
spatial inconsistency between unstructured point clouds presents challenges in propagating temporal information between frames.

To solve this problem,
motion information relay is proposed to initialize the temporal features
$c_{t-1}^{\prime}$ and $E_{t-1}^{\prime}$ embedded in ${PC}_{t}^L$ from $c_{t-1}$ and $E_{t-1}$ embedded in ${PC}_{t-1}^L$.
Based on the initial pose between two point clouds, we first warp the ($t$-$1$)-th frame of point cloud 
to roughly align two point clouds. Then, we utilize an MLP and max pooling to estimate $c_{t-1}^{\prime}$ and $E_{t-1}^{\prime}$:
{
\setlength{\abovedisplayskip}{2pt}
\setlength{\belowdisplayskip}{2pt}
\begin{align}
    \centering
    c_{t-1}^{\prime}&=\underset{k=1,2,\ldots,K}{M A X}\left(M L P\left( c_{t-1} \right)\right), \nonumber \\
    E_{t-1}^{\prime}&=\underset{k=1,2,\ldots,K}{M A X}\left(M L P\left( E_{t-1} \right)\right).
    \label{lstm2}
\end{align}
}

Then, a peephole LSTM~\cite{peephole1, peephole2} is adopted to propagate embedding features along time series.
A peephole LSTM comprises a cell state $c_t$ indicating long-term memory and a hidden state $h_t$ indicating short-term memory. We characterize the hidden state $h_t$ with the embedding feature $E_t$ to capture motion information in a short local sequence. 

Finally, the embedding feature $E_{t}$ and the cell state $c_{t}$ are sequentially propagated as follows:
{
\setlength{\abovedisplayskip}{2pt}
\setlength{\belowdisplayskip}{2pt}
\begin{align}
    x^{t} &= RE^{L} \oplus F^{L}, 
    \nonumber
\end{align}
\begin{align}
    f_{t} &= \sigma\left(MLP_f\left({c}_{t-1}^{\prime} \oplus E_{t-1}^{\prime} \oplus x_{t}\right)\right), 
    \nonumber
\end{align}
\begin{align}
    i_{t} &= \sigma\left(MLP_i\left({c}_{t-1}^{\prime} \oplus E_{t-1}^{\prime} \oplus x_{t}\right)\right),
    \nonumber
\end{align}
\begin{align}
    \tilde{c}_{t} &= \tanh \left(MLP_c\left(E_{t-1}^{\prime} \oplus x_{t}\right)\right), 
    \nonumber
\end{align}
\begin{align}
    c_{t} &= f_{t} \odot c_{t-1}^{\prime} + i_{t} \odot \tilde{c}_{t}, 
    \nonumber
\end{align}
\begin{align}
    o_{t} &= \sigma\left(MLP_o\left({c}_{t} \oplus E_{t-1}^{\prime} \oplus x_{t}\right)\right), 
    \nonumber
\end{align}
\begin{align}
    E_{t} &= o_{t} \odot \tanh \left(c_{t}\right),
    \label{ht}
\end{align}
}where $f_{t}$, $i_{t}$, and $o_{t}$ denote the activation weights of the forgetting gate, input gate, and output gate respectively. 

By propagating temporal information in LiDAR odometry, 
context and continuity in the time domain provide guidance for the pose estimation, resulting in enhanced robustness and accuracy of LiDAR odometry.


\begin{table*}[t]
    \caption{The LiDAR odometry experiment results on KITTI odometry dataset~\cite{KITTI1,KITTI2}. RTE and RRE mean the relative translation error ($\%$) and relative rotation error (${}^\circ /100m$) respectively, which are calculated by Root Mean Squared Error (RMSE) of the relative transformation on all possible subsequences in the length of $100,200,...,800\,m$. `$^*$' means the training sequence. 
    The best results are bold. The percentage increase in accuracy and reduction in runtime are calculated by comparing our DSLO with the baseline method PWCLO-Net~\cite{PWCLO}.}
    \footnotesize
    \centering
    \vspace{-5pt}
    \tabcolsep 0.65mm
    \renewcommand{\arraystretch}{1.1}
    \begin{tabular}{c||cc|cc|cc|cc|cc|cc|cc|cc||cc}
        \shline
        \multicolumn{1}{c||}{\multirow{2}{*}{Seq.}} & \multicolumn{2}{c|}{LOAM\cite{LOAM}} & \multicolumn{2}{c|}{ICP-po2po\cite{ICP}} & \multicolumn{2}{c|}{ICP-po2pl\cite{ICP}} & \multicolumn{2}{c|}{GICP\cite{GICP}} & \multicolumn{2}{c|}{CLS\cite{7487648}} & \multicolumn{2}{c|}{Velas et al.\cite{8374163}} & \multicolumn{2}{c|}{LO-Net\cite{LONET}} & \multicolumn{2}{c||}{PWCLO-Net\cite{PWCLO}} & \multicolumn{2}{c}{Ours} \\
        \cline{2-19}
        \multicolumn{1}{c||}{} &  RTE & RRE & RTE & RRE & RTE & RRE & RTE & RRE & RTE & RRE & RTE & RRE & RTE & RRE & RTE & RRE & RTE & RRE \\
        \shline
        $00^{*}$ & 1.10 & 0.53 & 6.88 & 2.99 & 3.80 & 1.73 & 1.29 & 0.64 & 2.11 & 0.95 & 3.02 & NA & 1.47 & 0.72 & \textbf{0.78} & 0.42 & \textbf{0.78} & \textbf{0.40} \\
        $01^{*}$ & 2.79 & 0.55 & 11.21 & 2.58 & 13.53 & 2.58 & 4.39 & 0.91 & 4.22 & 1.05 & 4.44 & NA & 1.36 & 0.47 & 0.67 & \textbf{0.23} & \textbf{0.66} & \textbf{0.23} \\
        $02^{*}$ & 1.54 & 0.55 & 8.21 & 3.39 & 9.00 & 2.74 & 2.53 & 0.77 & 2.29 & 0.86 & 3.42 & NA & 1.52 & 0.71 & 0.86 & 0.41 & \textbf{0.77} & \textbf{0.34} \\
        $03^{*}$ & 1.13 & 0.65 & 11.07 & 5.05 & 2.72 & 1.63 & 1.68 & 1.08 & 1.63 & 1.09 & 4.94 & NA & 1.03 & 0.66 & 0.76 & 0.44 & \textbf{0.67} & \textbf{0.37} \\ 
        $04^{*}$ & 1.45 & 0.50 & 6.64 & 4.02 & 2.96 & 2.58 & 3.76 & 1.07 & 1.59 & 0.71 & 1.77 & NA & 0.51 & 0.65 & 0.37 & \textbf{0.40} & \textbf{0.31} & 0.47 \\
        $05^{*}$ & 0.75 & 0.38 & 3.97 & 1.93 & 2.29 & 1.08 & 1.02 & 0.54 & 1.98 & 0.92 & 2.35 & NA & 1.04 & 0.69 & \textbf{0.45} & \textbf{0.27} & 0.50 & 0.30 \\ 
        $06^{*}$ & 0.72 & 0.39 & 1.95 & 1.59 & 1.77 & 1.00 & 0.92 & 0.46 & 0.92 & 0.46 & 1.88 & NA & 0.71 & 0.50 & \textbf{0.27} & \textbf{0.22} & 0.57 & 0.38 \\
        $07$ & 0.69 & 0.50 & 5.17 & 3.35 & 1.55 & 1.42 & 0.64 & 0.45 & 1.04 & 0.73 & 1.77 & NA & 1.70 & 0.89 & 0.60 & 0.44 & \textbf{0.58} & \textbf{0.41} \\
        $08$ & 1.18 & 0.44 & 10.04 & 4.93 & 4.42 & 2.14 & 1.58 & 0.75 & 2.14 & 1.05 & 2.89 & NA & 2.12 & 0.77 & 1.26 & 0.55 & \textbf{1.16} & \textbf{0.51} \\
        $09$ & 1.20 & 0.48 & 6.93 & 2.89 & 3.95 & 1.71 & 1.97 & 0.77 & 1.95 & 0.92 & 4.94 & NA & 1.37 & 0.58 & 0.79 & 0.35 & \textbf{0.72} & \textbf{0.33} \\
        $10$ & 1.51 & 0.57 & 8.91 & 4.74 & 6.13 & 2.60 & 1.31 & 0.62 & 3.46 & 1.28 & 3.27 & NA & 1.80 & 0.93 & 1.69 & 0.62 & \textbf{1.29} & \textbf{0.49} \\	
        \hline 
        \multirow{2}{*}{\parbox{1.0cm}{Mean on 07-10}} & \multirow{2}{*}{1.145} & \multirow{2}{*}{0.498} & \multirow{2}{*}{7.763} & \multirow{2}{*}{3.978} & \multirow{2}{*}{4.013} & \multirow{2}{*}{1.968} & \multirow{2}{*}{1.375} & \multirow{2}{*}{0.648} & \multirow{2}{*}{2.148} & \multirow{2}{*}{0.995} & \multirow{2}{*}{3.218} & \multirow{2}{*}{NA} & \multirow{2}{*}{1.748} & \multirow{2}{*}{0.793} & \multirow{2}{*}{1.085} & \multirow{2}{*}{0.490} & \textbf{0.938} & \textbf{0.435} \\
        & & & & & & & & & & & & & & & & & \textcolor{red}{(\textbf{$\uparrow$15.67\%})} & \textcolor{red}{(\textbf{$\uparrow$12.64\%})} \\
        \shline 
        Time(s) & \multicolumn{2}{c|}{0.069} & \multicolumn{2}{c|}{0.61} & \multicolumn{2}{c|}{0.98} & \multicolumn{2}{c|}{3.21} & \multicolumn{2}{c|}{2.36} & \multicolumn{2}{c|}{0.067} & \multicolumn{2}{c|}{0.08} & \multicolumn{2}{c||}{0.066} & \multicolumn{2}{c}{\textbf{0.049} \textcolor{red}{(\textbf{$\downarrow$ 34.69\%})}} \\ 
        \shline
    \end{tabular}
    \vspace{-10pt}
    \label{tab:result_transposed}
\end{table*}

\subsection{Training Loss}\label{sec:loss}
During the training process, three frames of point clouds are processed simultaneously. Then the estimated poses between each two frames
are integrated into the loss $\ell$:
{
\setlength{\abovedisplayskip}{2pt}
\setlength{\belowdisplayskip}{2pt}
\begin{equation}
    \ell=\sum_{l=1}^{L} \alpha^{l} \ell^{l}_{t,t+1} + \sum_{l=1}^{L} \alpha^{l} \ell^{l}_{t+1,t+2} + \sum_{l=1}^{L} \alpha^{l} \ell^{l}_{t,t+2},
    \label{l1}  
\end{equation}
}where each loss term
is determined by a weighted sum of losses from different layers of point clouds. Here, $\alpha^l$ denotes the weights at the $l$-th layer.

For the loss at the $l$-th layer, we adopt the design proposed in~\cite{LONET, PWCLO, wang2022efficient}:
{
\setlength{\abovedisplayskip}{4pt}
\setlength{\belowdisplayskip}{4pt}
\begin{equation}
    \begin{aligned}
        \ell^{l}_{t,t+1} &=\left\|t_{g t}-t^{l}\right\| \exp \left(-s_{t}\right)+s_{t} \\
        &+\left\|q_{g t}-\frac{q^{l}}{\left\|q^{l}\right\|}\right\|_{2} \exp \left(-s_{q}\right)+s_{q},
    \end{aligned}
    \label{l2}
\end{equation}
}where $\|\cdot\|$ and $\|\cdot\|_2$ denote the $\ell_1$ and $\ell_2$ norm respectively.
$s_t$ and $s_q$ are two learnable parameters designed to balance the scale differences in translation and rotation errors of the estimated poses.
$t^{l}$ and $q^{l}$ represent the pose predicted
at the $l$-th layer,
while $t_{gt}$ and $q_{gt}$ denote the ground truth.

\section{Experiments}


\subsection{Implementation Details}
\label{sec:Implementation}

The proposed network is implemented with PyTorch and trained/tested on an NVIDIA RTX2080Ti GPU and an Intel Xeon W-2265 3.50GHz CPU. The initial learning rate is 0.001, decaying by 0.7 every 26 epochs until it reaches 0.00001. We use the Adam optimizer with $\beta_1=0.9$ and $\beta_2=0.999$, a batch size of 8, and an input point cloud size of $N=8192$. A dropout layer is included during pose regression to mitigate overfitting.

In Equation~\eqref{l1}, $L=4$, $\alpha^1=1.6$, $\alpha^2=0.8$, $\alpha^3=0.4$, $\alpha^4=0.2$. The initial values for learnable parameters in the loss function are $s_t=0$ and $s_q=-2.5$ in Equation~\eqref{l2}.

During training, three frames of point cloud are processed simultaneously. During testing, poses are continuously estimated to maintain motion information across the sequence.

\subsection{Accuracy Evaluation}

\begin{figure}[t]
	\centering
	\subfloat[Seq. 07]{\includegraphics[width=.37\columnwidth]{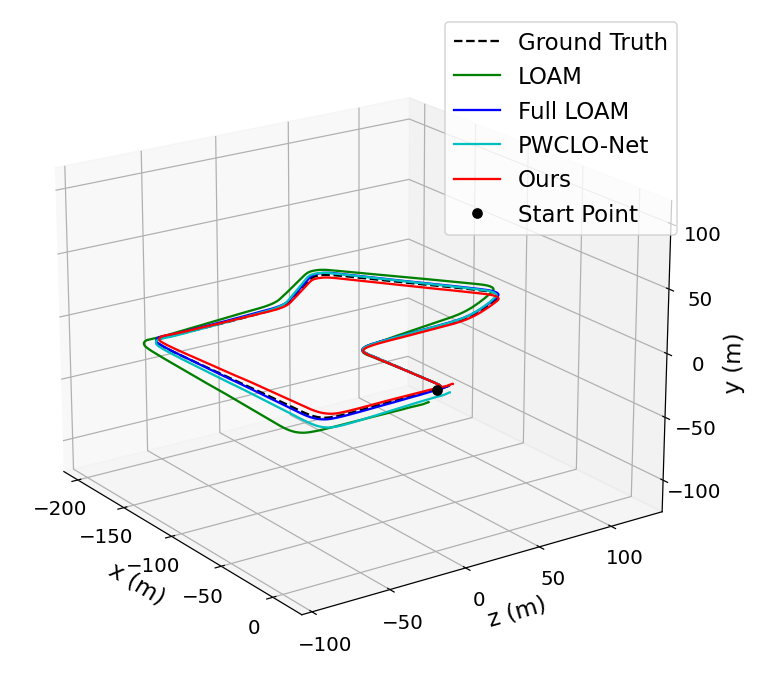}}\hspace{15pt}
     \vspace{-5pt}
	\subfloat[Seq. 08]
    {\includegraphics[width=.37\columnwidth]{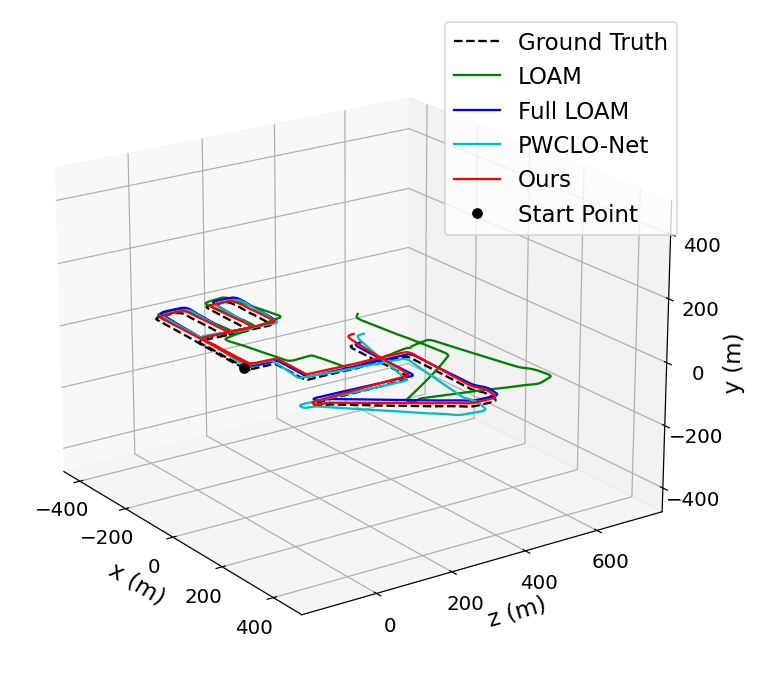}}\\
     \vspace{-5pt}
	\subfloat[Seq. 09]{\includegraphics[width=.37\columnwidth]{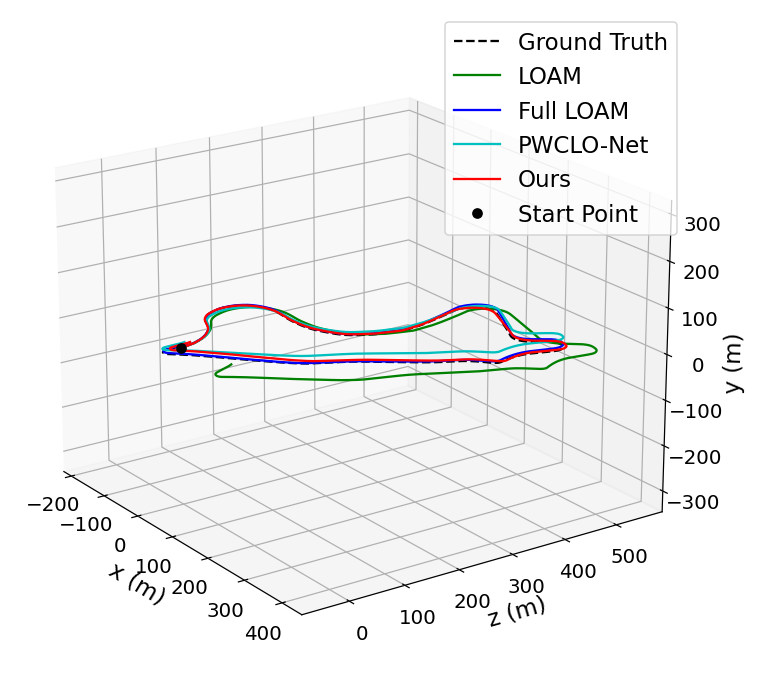}}\hspace{15pt}
	\subfloat[Seq. 10]{\includegraphics[width=.37\columnwidth]{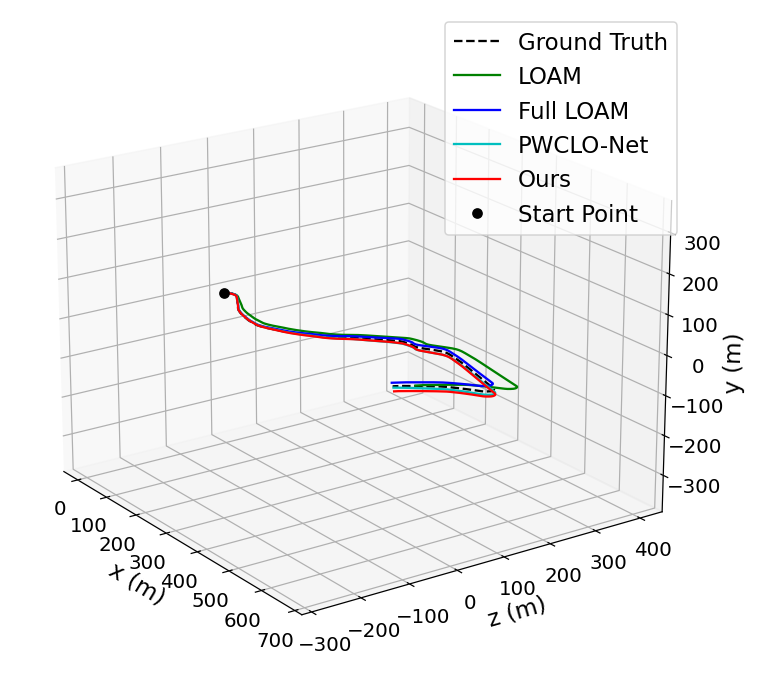}}
	\caption{3D trajectory results on KITTI Seq. 07-10.}
	\vspace{-10pt}
    \label{fig:kitti_vis}
\end{figure}

\begin{table}[t]
    \begin{center}
        \caption{Odometry experiments on
        Argoverse dataset~\cite{argo}. 
        The best results are bold.}
        \footnotesize
        \vspace{-5pt}
        \tabcolsep 1pt
        \renewcommand{\arraystretch}{1.2}
        \begin{tabular}{p{6.3cm}||cc}
            \shline
            \multirow{2}{*}{$\text {Method}$} & 
            \multicolumn{2}{c}{$\text { Mean on 00-23 }$} \\
            \cline{2-3} & ATE & RPE \\
            \shline
            \text { LeGO-LOAM~\cite{LeGO-LOAM} w/o mapping } & 4.537 & 0.110 \\
            \hline \text { SUMA~\cite{suma} w/o mapping } & 3.663 & 0.039 \\
            \hline \text { PyLiDAR~\cite{pylidar} w/o mapping } & 6.900 & 0.109 \\
            \hline \text { A-LOAM~\cite{ALOAM} w/o mapping } & 4.138 & 0.066 \\
            \hline \hline \text { Ours } & \textbf{0.111}	& \textbf{0.027} \\
            \shline
        \end{tabular}
        \vspace{-15pt}
        \label{tab:argo-acc}    
    \end{center}
\end{table}

\begin{table}[t]
    \footnotesize
	\begin{center}
        \caption{Runtime of DSLO and other LiDAR odometry. 
        }
        \vspace{-5pt}
        \tabcolsep 1pt
        \renewcommand{\arraystretch}{1.1}
		\begin{tabular}{l||c}
            \shline
            Method                   & Runtime/s     \\ 
            \shline
            ICP-po2po~\cite{ICP}        & 0.61           \\ 
            ICP-po2pl~\cite{ICP}        & 0.98           \\ 
            GICP~\cite{GICP}        & 3.21           \\ 
            LOAM~\cite{LOAM}            & 0.069            \\ 
            CLS~\cite{7487648}          & 2.36              \\ 
            Velas et al.~\cite{8374163}             &     0.067        \\ 
            LO-Net~\cite{LONET}          & 0.080              \\ 
            PWCLO-Net~\cite{PWCLO}          & 0.066              \\ 
            \hline \hline 
            Ours w/o pose initialization \& pyramid feature sharing & 0.072          \\ 
            Ours w/o pose initialization & 0.056          \\ 
            Ours(full, with pose initialization \& pyramid feature sharing)              & \textbf{0.049} \\ 
            \shline
		\end{tabular}
		\vspace{-20pt}
		\label{tab:runtime}
	\end{center}
\end{table}

\subsubsection{Experiments on KITTI odometry dataset}


We use KITTI Seq. 00-06 for training and 07-10 for validation.
The quantitative results are compared with recent LiDAR odometry methods~\cite{LOAM, GICP, LONET, PWCLO,ICP, 7487648, 8374163} in Table~\ref{tab:result_transposed}.
The original ICP-po2po~\cite{ICP} 
performs poorly due to improper initial values,
while its variants ICP-po2pl~\cite{ICP} and GICP~\cite{GICP} perform better but lack robustness, with large localization errors on certain sequences.
Feature-based methods like LOAM~\cite{LOAM} and CLS~\cite{7487648} use hand-crafted features, whereas our DSLO uses a learning-based feature extraction strategy, achieving superior accuracy. 
Unlike other learning-based LiDAR odometry methods, such as LO-Net~\cite{LONET} and Velas et al.~\cite{8374163}, which project LiDAR data onto 2D planes, our DSLO processes raw point clouds directly. This enables it to learn more comprehensive 3D spatial structure information.
DSLO also exhibits accuracy improvements over 
the baseline method PWCLO-Net~\cite{PWCLO}. 
PWCLO-Net estimates pose between two point clouds independently.
In contrast, our network ensures motion consistency through sequential pose initialization and temporal feature propagation. 
Overall, our DSLO outperforms all recent LiDAR odometry methods with at least a 15.67\% improvement on RTE and a 12.64\% improvement on RRE
on average evaluation.

Qualitative results in Fig.~\ref{fig:kitti_vis}
show the trajectory estimated by our DSLO coincides with the ground truth best. 

\begin{table*}[t]
    \footnotesize
    \centering
    \caption{The ablation study results of DSLO for the network structure on KITTI odometry dataset~\cite{KITTI1,KITTI2}.}
    \vspace{-5pt}
    \tabcolsep 0.45mm
    \renewcommand{\arraystretch}{1.1}
    \begin{tabular}{l||p{9.8cm}||cc|cc|cc|cc||cc}
        \shline
        \multirow{2}{*}{}    & \multirow{2}{*}{Method}   & \multicolumn{2}{c|}{07}                                              & \multicolumn{2}{c|}{08}                                              & \multicolumn{2}{c|}{09}                                              & \multicolumn{2}{c||}{10}                                              & \multicolumn{2}{c}{Mean on 07-10}                                    \\ \cline{3-12} 
                 &                           & \multicolumn{1}{c}{RTE}    & \multicolumn{1}{c|}{RRE} & \multicolumn{1}{c}{RTE}    & \multicolumn{1}{c|}{RRE} & \multicolumn{1}{c}{RTE}    & \multicolumn{1}{c|}{RRE} & \multicolumn{1}{c}{RTE}    & \multicolumn{1}{c||}{RRE} & \multicolumn{1}{c}{RTE}     & \multicolumn{1}{c}{RRE} \\ 
                 \shline
\multirow{3}{*}{(a)} & Ours, w/o sequential pose initialization & \multicolumn{1}{c}{0.60} & 0.49 & \multicolumn{1}{c}{1.33} & 0.49 & \multicolumn{1}{c}{0.99} & 0.39 & \multicolumn{1}{c}{\textbf{1.20}} & 0.62 & \multicolumn{1}{c}{1.029} & 0.496 \\  
                 & Ours, w/o sequential pose initialization, w/o spatial information reuse           & \multicolumn{1}{c}{\textbf{0.58}} & 0.50                   & \multicolumn{1}{c}{1.28}          & \textbf{0.42}                   & \multicolumn{1}{c}{0.85}          & 0.38                            & \multicolumn{1}{c}{1.44}          & 0.58                            & \multicolumn{1}{c}{1.039}          & 0.470                           \\ 
                 & Ours (full, with sequential pose initialization \& spatial information reuse)       &        \textbf{0.58}	& \textbf{0.41}	& \textbf{1.16}	& 0.51	& \textbf{0.72}	& \textbf{0.33}	& 1.29	& \textbf{0.49}	& \textbf{0.938}	& \textbf{0.435}
                  \\ \hline
\multirow{2}{*}{(b)} & Ours, replace gated hierarchical pose refinement with MLP  & \multicolumn{1}{c}{1.05}          & 0.85                            & \multicolumn{1}{c}{1.40}          & 0.60                            & \multicolumn{1}{c}{0.90}          & 0.43                            & \multicolumn{1}{c}{\textbf{1.20}} & 0.53                            & \multicolumn{1}{c}{1.136}          & 0.603                           \\ 
                 & Ours (full, with gated hierarchical pose refinement) & \textbf{0.58}	& \textbf{0.41}	& \textbf{1.16}	& \textbf{0.51}	& \textbf{0.72}	& \textbf{0.33}	& 1.29	& \textbf{0.49}	& \textbf{0.938}	& \textbf{0.435}\\ \hline
\multirow{3}{*}{(c)} & Ours, w/o motion information relay, w/o peephole LSTM                      & \multicolumn{1}{c}{0.73}          & 0.53                            & \multicolumn{1}{c}{1.29}          & \textbf{0.46}                          & \multicolumn{1}{c}{0.93}          & 0.43                            & \multicolumn{1}{c}{\textbf{1.22}}          & 0.57                          & \multicolumn{1}{c}{1.041}          & 0.498                          \\ 
                 & Ours, with nearest neighbor query \& peephole LSTM & \multicolumn{1}{c}{1.50}          & 0.76                            & \multicolumn{1}{c}{1.53}          & 0.54                            & \multicolumn{1}{c}{0.95}          & 0.45                            & \multicolumn{1}{c}{0.88} & 0.58                            & \multicolumn{1}{c}{1.220}          & 0.580                           \\ 
                 & Ours (full, with motion information relay \& peephole LSTM) & \textbf{0.58}	& \textbf{0.41}	& \textbf{1.16}	& 0.51	& \textbf{0.72}	& \textbf{0.33}	& 1.29	& \textbf{0.49}	& \textbf{0.938}	& \textbf{0.435}\\ 
                 \shline
    \end{tabular}
    \vspace{-10pt}
    \label{tab:Ablation}
\end{table*}

\subsubsection{Experiments on Argoverse dataset}


To assess the generalization of our method, we conduct experiments on the Argoverse dataset~\cite{argo}.
We employ Absolute Trajectory Error (ATE) and Relative Pose Error (RPE) for evaluation due to the short sequence length. 
Our DSLO model is trained and tested on the official Argoverse training/testing split. 
For comparison, we assess four geometry-based odometry methods~\cite{LeGO-LOAM, suma, pylidar, ALOAM} with the mapping thread disabled. 
As shown in Table~\ref{tab:argo-acc}, our DSLO significantly outperforms these methods, particularly in the ATE metric.

\subsection{Real-time Performance Evaluation}
To validate the real-time performance of the proposed method, we evaluate the runtime of DSLO, DSLO with efficiency modules removed, and the baseline method PWCLO-Net~\cite{PWCLO} on KITTI Seq. 04 with a batch size of 1. 
The experimental results in Table~\ref{tab:runtime} demonstrate that our sequential pose initialization and spatial information reuse modules reduce runtime by 34.69\% compared to PWCLO-Net. 
Furthermore, DSLO achieves a comparable inference speed with other registration-based~\cite{GICP,ICP}, feature extraction-based~\cite{LOAM,7487648}, and learning-based methods~\cite{LONET,8374163}.


\subsection{Ablation Study}

\subsubsection{Spatial information reuse and sequential pose initialization} 

We first eliminate sequential pose initialization, using attentive cost volume for initial pose prediction across the sequence. Next, we remove spatial information reuse, resulting in redundant calculations of the same point feature pyramid. As shown in Table~\ref{tab:Ablation} (a), both techniques improve accuracy and enhance real-time performance (Table~\ref{tab:runtime}).

\subsubsection{Gated hierarchical pose refinement} 

We replace the gated hierarchical pose refinement module with an MLP structure. Results in Table~\ref{tab:Ablation} (b) indicate that the gated module outperforms the MLP in refining poses. The gated hierarchical pose refinement module is adept at preserving hierarchical information through self-learning gate states. It effectively filters valid information from different layers.

\subsubsection{Temporal information propagation with inconsistent spatial context} 
We first remove the entire temporal feature propagation module, including motion information relay and peephole LSTM. 
Subsequently, 
the learning-based motion information relay method is replaced with a direct nearest neighbor query.
Specifically, the LSTM state value of the nearest point in the ($t$-$1$)-th frame is assigned to the corresponding point in the $t$-th frame.
Results in Table~\ref{tab:Ablation} (c) demonstrate the significance of the proposed temporal feature propagation method, particularly the learning-based motion information relay, 
in improving pose regression accuracy.


\section{Conclusion}


We propose a novel deep sequence LiDAR odometry based on inconsistent spatio-temporal propagation, achieving high accuracy and real-time performance.
Our spatial information reuse and sequential pose initialization reduce the computational overhead without degrading the accuracy.  
The gated hierarchical pose refinement enables efficient coarse-to-fine pose updates, with self-learning gate estimations discarding incorrect historical matches and mitigating sensor noise. Additionally, our temporal feature propagation fuses motion information over time, addressing inconsistent spatial context and improving LiDAR odometry accuracy.

\normalem
\bibliographystyle{IEEEtran}  
\bibliography{IEEEabrv,root} 

\begin{thebibliography}{10}
\providecommand{\url}[1]{#1}
\csname url@rmstyle\endcsname
\providecommand{\newblock}{\relax}
\providecommand{\bibinfo}[2]{#2}
\providecommand\BIBentrySTDinterwordspacing{\spaceskip=0pt\relax}
\providecommand\BIBentryALTinterwordstretchfactor{4}
\providecommand\BIBentryALTinterwordspacing{\spaceskip=\fontdimen2\font plus
\BIBentryALTinterwordstretchfactor\fontdimen3\font minus
  \fontdimen4\font\relax}
\providecommand\BIBforeignlanguage[2]{{%
\expandafter\ifx\csname l@#1\endcsname\relax
\typeout{** WARNING: IEEEtran.bst: No hyphenation pattern has been}%
\typeout{** loaded for the language `#1'. Using the pattern for}%
\typeout{** the default language instead.}%
\else
\language=\csname l@#1\endcsname
\fi
#2}}

\bibitem{li2023hong}
H.~Li, J.~Zhao, J.-C. Bazin, P.~Kim, K.~Joo, Z.~Zhao, and Y.-H. Liu, ``Hong
  kong world: Leveraging structural regularity for line-based slam,''
  \emph{TPAMI}, vol.~45, no.~11, pp. 13\,035--13\,053, 2023.

\bibitem{li2020robust}
H.~Li, J.~Zhao, J.-C. Bazin, and Y.-H. Liu, ``Robust estimation of absolute
  camera pose via intersection constraint and flow consensus,'' \emph{TIP},
  vol.~29, pp. 6615--6629, 2020.

\bibitem{LOAM}
J.~Zhang and S.~Singh, ``{LOAM:} lidar odometry and mapping in real-time,'' in
  \emph{RSS}, vol.~2, no.~9, 2014, pp. 1--9.

\bibitem{GICP}
A.~Segal, D.~H{\"{a}}hnel, and S.~Thrun, ``Generalized-icp,'' in \emph{RSS},
  vol.~2, no.~4, 2009, p. 435.

\bibitem{LeGO-LOAM}
T.~Shan and B.~Englot, ``Lego-loam: Lightweight and ground-optimized lidar
  odometry and mapping on variable terrain,'' in \emph{IROS}, 2018, pp.
  4758--4765.

\bibitem{suma}
J.~Behley and C.~Stachniss, ``Efficient surfel-based slam using 3d laser range
  data in urban environments,'' in \emph{RSS}, vol. 2018, 2018, p.~59.

\bibitem{LodoNet}
C.~Zheng, Y.~Lyu, M.~Li, and Z.~Zhang, ``Lodonet: A deep neural network with 2d
  keypoint matching for 3d lidar odometry estimation,'' in \emph{ASE}, 2020,
  pp. 2391--2399.

\bibitem{LONET}
Q.~Li, S.~Chen, C.~Wang, X.~Li, C.~Wen, M.~Cheng, and J.~Li, ``Lo-net: Deep
  real-time lidar odometry,'' in \emph{CVPR}, 2019, pp. 8465--8474.

\bibitem{liu2023translo}
J.~Liu, G.~Wang, C.~Jiang, Z.~Liu, and H.~Wang, ``Translo: A window-based
  masked point transformer framework for large-scale lidar odometry,'' in
  \emph{AAAI}, vol.~37, no.~2, 2023, pp. 1683--1691.

\bibitem{liu2023regformer}
J.~Liu, G.~Wang, Z.~Liu, C.~Jiang, M.~Pollefeys, and H.~Wang, ``Regformer: An
  efficient projection-aware transformer network for large-scale point cloud
  registration,'' in \emph{ICCV}, 2023, pp. 8451--8460.

\bibitem{PWCLO}
G.~Wang, X.~Wu, Z.~Liu, and H.~Wang, ``Pwclo-net: Deep lidar odometry in 3d
  point clouds using hierarchical embedding mask optimization,'' in
  \emph{CVPR}, 2021, pp. 15\,905--15\,914.

\bibitem{zhou2023hpplo}
B.~Zhou, Y.~Tu, Z.~Jin, C.~Xu, and H.~Kong, ``Hpplo-net: Unsupervised lidar
  odometry using a hierarchical point-to-plane solver,'' \emph{IEEE TIV},
  vol.~9, no.~1, pp. 2727--2739, 2023.

\bibitem{KITTI1}
A.~Geiger, P.~Lenz, and R.~Urtasun, ``Are we ready for autonomous driving? the
  kitti vision benchmark suite,'' in \emph{CVPR}, 2012, pp. 3354--3361.

\bibitem{KITTI2}
A.~Geiger, P.~Lenz, C.~Stiller, and R.~Urtasun, ``Vision meets robotics: The
  kitti dataset,'' \emph{IJRR}, vol.~32, no.~11, pp. 1231--1237, 2013.

\bibitem{argo}
M.-F. Chang, J.~Lambert, P.~Sangkloy, J.~Singh, S.~Bak, A.~Hartnett, D.~Wang,
  P.~Carr, S.~Lucey, D.~Ramanan, \emph{et~al.}, ``Argoverse: 3d tracking and
  forecasting with rich maps,'' in \emph{CVPR}, 2019.

\bibitem{DMLO}
Z.~Li and N.~Wang, ``Dmlo: Deep matching lidar odometry,'' in \emph{IROS},
  2020, pp. 6010--6017.

\bibitem{DeepCLR}
M.~Horn, N.~Engel, V.~Belagiannis, M.~Buchholz, and K.~Dietmayer, ``Deepclr:
  Correspondence-less architecture for deep end-to-end point cloud
  registration,'' in \emph{ITSC}, 2020, pp. 1--7.

\bibitem{wang2021anchor}
G.~Wang, H.~Liu, M.~Chen, Y.~Yang, Z.~Liu, and H.~Wang, ``Anchor-based
  spatio-temporal attention 3-d convolutional networks for dynamic 3-d point
  cloud sequences,'' \emph{IEEE TIM}, vol.~70, pp. 1--11, 2021.

\bibitem{wang2021self}
H.~Wang, L.~Yang, X.~Rong, J.~Feng, and Y.~Tian, ``Self-supervised 4d
  spatio-temporal feature learning via order prediction of sequential point
  cloud clips,'' in \emph{WACV}, 2021, pp. 3762--3771.

\bibitem{fan2021point}
H.~Fan, Y.~Yang, and M.~Kankanhalli, ``Point 4d transformer networks for
  spatio-temporal modeling in point cloud videos,'' in \emph{CVPR}, 2021, pp.
  14\,204--14\,213.

\bibitem{wang2022sti}
Y.~Wang, H.~Pan, J.~Zhu, Y.-H. Wu, X.~Zhan, K.~Jiang, and D.~Yang, ``Be-sti:
  Spatial-temporal integrated network for class-agnostic motion prediction with
  bidirectional enhancement,'' in \emph{CVPR}, 2022, pp. 17\,093--17\,102.

\bibitem{POINTNET2}
C.~R. Qi, L.~Yi, H.~Su, and L.~J. Guibas, ``Pointnet++: Deep hierarchical
  feature learning on point sets in a metric space,'' in \emph{NIPS}, 2017, p.
  5105–5114.

\bibitem{FLOWNET}
X.~Liu, C.~R. Qi, and L.~J. Guibas, ``Flownet3d: Learning scene flow in 3d
  point clouds,'' in \emph{CVPR}, 2019, pp. 529--537.

\bibitem{FLOWSTEP}
Y.~Kittenplon, Y.~C. Eldar, and D.~Raviv, ``Flowstep3d: Model unrolling for
  self-supervised scene flow estimation,'' in \emph{CVPR}, 2021.

\bibitem{RAFT}
Z.~Teed and J.~Deng, ``Raft: Recurrent all-pairs field transforms for optical
  flow,'' in \emph{ECCV}, 2020, pp. 402--419.

\bibitem{peephole1}
F.~Gers and E.~Schmidhuber, ``Lstm recurrent networks learn simple context-free
  and context-sensitive languages,'' \emph{IEEE TNN}, vol.~12, no.~6, pp.
  1333--1340, 2001.

\bibitem{peephole2}
F.~A. Gers, N.~N. Schraudolph, and J.~Schmidhuber, ``Learning precise timing
  with lstm recurrent networks,'' \emph{JMLR}, vol.~3, no.~1, pp. 115--143,
  2003.

\bibitem{ICP}
P.~J. Besl and N.~D. McKay, ``A method for registration of 3-d shapes,''
  \emph{TPAMI}, vol.~14, no.~2, pp. 239--256, 1992.

\bibitem{7487648}
M.~Velas, M.~Spanel, and A.~Herout, ``Collar line segments for fast odometry
  estimation from velodyne point clouds,'' in \emph{ICRA}, 2016, pp.
  4486--4495.

\bibitem{8374163}
M.~Velas, M.~Spanel, M.~Hradis, and A.~Herout, ``Cnn for imu assisted odometry
  estimation using velodyne lidar,'' in \emph{ICARSC}, 2018.

\bibitem{wang2022efficient}
G.~Wang, X.~Wu, S.~Jiang, Z.~Liu, and H.~Wang, ``Efficient 3d deep lidar
  odometry,'' \emph{TPAMI}, vol.~45, no.~5, pp. 5749--5765, 2022.

\bibitem{pylidar}
P.~Dellenbach, J.-E. Deschaud, B.~Jacquet, and F.~Goulette, ``What’s in my
  lidar odometry toolbox?'' in \emph{IROS}, 2021, pp. 4429--4436.

\bibitem{ALOAM}
\BIBentryALTinterwordspacing
T.~Qin and S.~Cao. {A-LOAM:} advanced implementation of loam. [Online].
  Available: \url{https://github.com/HKUST-Aerial-Robotics/A-LOAM}
\BIBentrySTDinterwordspacing

\end{thebibliography}

\end{document}